\PassOptionsToPackage{table}{xcolor}
\documentclass[11pt]{article}

\usepackage[final]{acl}

\usepackage{booktabs}
\usepackage{tabularx}
\usepackage{float}
\usepackage{dblfloatfix}

\usepackage{times}
\usepackage{latexsym}

\usepackage[T1]{fontenc}

\usepackage[utf8]{inputenc}

\usepackage{microtype}

\usepackage{inconsolata}

\usepackage{graphicx}

\title{Narrative Landscape: Mapping Narrative Dispositions Across LLMs}

\author{Donghoon Jung \quad
  Jiwoo Choi\thanks{Co-second authors.} \quad
  Songeun Chae\footnotemark[1] \quad
  Seohyon Jung\thanks{Corresponding author.} \\
  School of Digital Humanities and Computational Social Sciences, KAIST, South Korea \\
  \texttt{\{donghoon.jung, jwchoi0515, songeun, seohyon.jung\}@kaist.ac.kr}}

\begin{document}
\maketitle
\begin{abstract}
This study proposes a quantitative framework for profiling LLM dispositions as stable, model-specific regularities in output under repeated, controlled elicitation. Using a structured narrative constraint-selection task administered across six frontier models and three instruction types, we operationalize disposition through two dimensions: “consistency”, measured as cross-replication selection overlap via Jaccard similarity, and “diversity”, measured as dispersion across options via the inverse Simpson index. We further introduce Narrative Landscape, a PCA-based visualization that maps each model’s selection profile into a shared space for direct comparison. Results reveal a clear rigidity--exploration spectrum across model families and show that instruction types shift the geometry of selection spaces even when scalar metrics appear similar, indicating that comparable scores can mask qualitatively distinct selection topologies.
\end{abstract}

\section{Introduction}

During the development of Claude Opus 4.5, researchers reportedly used an internal “soul document” to specify the model’s persona, values, and interaction style \citep{weiss2025soulDocument}. \citet{anthropic2024claudeCharacter} characterizes this broader practice as character training, which aims to instill more nuanced traits such as curiosity and open-mindedness into the model’s responses through design choices beyond standard instruction following. As large language models (LLMs) are now widely used across domains including narrative writing, scientific writing, and ideation, their engineered dispositions become consequential for the kinds of outputs they systematically produce. This study argues for the need to identify such dispositions as empirically observable regularities.

Recent empirical work uses replication-based elicitation to probe disposition stability across personality inventories \citep{serapioGarcia2023personalityTraitsResearchSquare}, political preference, and moral robustness and susceptibility under persona role-play \citep{rozado2024politicalPreferences, costa2025moralSusceptibility}, and large-scale comparisons of personality versus political profiles \citep{goyanes2025personalityPoliticalDispositions}, alongside measurement refinements via open-ended, AI-rated Big Five assessment \citep{zheng-etal-2025-lmlpa}. Yet such profiles remain fragile with respect to test--retest reliability and paraphrase variation \citep{wang2025comparativeLLMHumanPersonality} and exhibit limited behavioral coherence, including word--deed inconsistency and weak cross-aspect transfer \citep{xu2025sayOneThingDoAnother}; moreover, they can be stabilized at the level of self-report without commensurate behavioral change \citep{han2025personalityIllusion}. Despite these measurement advances, prior work lacks a shared-space visualization that exposes model- and instruction-level structure in repeated selections.

This study introduces a replication-based framework for profiling LLM disposition as model-specific regularities in narrative constraint selection under controlled elicitation \citep{jung-etal-2025-style}. With a fixed pool of narrative constraints and repeated replications, overlap in selected constraints captures the stability of commitments, operationalized as consistency via Jaccard similarity \citep{broder1997resemblanceContainment}, while dispersion of selections across the pool captures coverage of the decision space, operationalized as diversity via the inverse Simpson index \citep{hill1973diversityEvenness}. To complement these scalar summaries and expose differences in selection structure, we also embed constraint-frequency profiles into a shared Principal Component Analysis (PCA) space for direct comparison across models and instruction types.

Results show a clear spectrum of narrative dispositions across model families, from rigid, high-overlap constraint selections with limited option-space coverage to low-overlap selections with markedly higher diversity; instruction types further modulate the extent of the selection space. Narrative Landscape also indicates that comparable scalar metrics can correspond to distinct selection topologies, representing model- and instruction-level differences as geometric structure in a shared constraint space. The measurement framework and Narrative Landscape generalize beyond narrative, providing a transferable approach to profiling disposition in structured selection tasks.

\section{Data}
\paragraph{Model Selection and Constraint Pool.}
Building on \citet{jung-etal-2025-style}, we analyze six state-of-the-art LLMs spanning major provider families (OpenAI, Anthropic, Google, Alibaba). To maximize replicability across repeated runs, decoding parameters were held constant where available; temperature (temp) and top-p were fixed at 1.0 for all models, and vendor-specific controls (reasoning effort, verbosity) were set to high when present. Model identifiers and decoding settings are summarized in \autoref{app:models-decoding}.

All models were presented with the same constraint pool, which functions as a theory-grounded diagnostic taxonomy for making narrative preferences observable as authorial choices. The pool contains 200 narrative constraints systematically distributed across four narratological elements—Event, Style, Character, Setting—with each element subdivided into five categories of ten constraints each. To minimize surface-level selection bias and ensure that observed differences reflect model dispositions rather than prompt artifacts, each constraint is designed with structural regularity (uniform word length, parallel grammatical structure) and matched conceptual granularity within categories. The models operate on the natural-language constraints themselves, allowing selection patterns to be interpreted as behavioral traces of a model’s implicit narrative logic.

\paragraph{Experimental Procedure.}
Each observation in our dataset corresponds to a single run in which a model produces a narrative plan by selecting constraints and generating justifications under a controlled instruction type. We operationalize “authorial orientations” through three broad instruction types—Basic, Quality-focused, and Creativity-focused—intended to capture stance rather than dependence on specific phrasing. Across runs, the persona is set via the instruction, and the user prompt instructs the model to read the full constraint pool, select a fixed number of constraints, justify each selection, and then evaluate cross-constraint dynamics in a final compatibility assessment.

In this paper, we focus on a pooled, unlabeled constraint-selection setting with a fixed selection budget. Models are presented with the full pool of 200 constraints without element labels and are required to select exactly 20 constraints they deem most useful for planning a single fictional narrative, providing a brief justification for each selection. The full prompt templates and the complete constraint pool are documented in \citet{jung-etal-2025-style}.

To enable rigorous replication and to control for order effects \citep{liu-etal-2024-lost, pezeshkpour-hruschka-2024-large, shi2025judgingTheJudges}, each run uses (i) a fresh random permutation of the constraint list and (ii) an isolated session state. Within any fixed experimental cell (model × instruction type, in our setting), only the constraint-order permutation and provider stochasticity vary; all other factors are held constant. We execute 160 independent replications per cell (2,880 runs total), producing dense samples for stability and sensitivity analyses.

\section{Consistency and Diversity}
To measure the consistency of constraint selection across repeated runs within the same model--instruction type cell, we use Jaccard similarity ($J$) \citep{broder1997resemblanceContainment}:
\[
J(A,B)=\frac{|A\cap B|}{|A\cup B|}
\]
Here, $A$ and $B$ are the constraint sets selected in two runs. For each model--instruction type cell, we compute Jaccard similarity over all run pairs and report the mean pairwise value as the consistency score.

To measure diversity, we use the Gini--Simpson index ($GS$) \citep{simpson1949measurementOfDiversity}, which corresponds to the probability that two randomly selected constraints are different. Because the Gini--Simpson index is bounded in $[0,1]$ and can compress differences, we also report the inverse Simpson index in its effective-number form ($EN$) \citep{hill1973diversityEvenness}:
\[
GS = 1 - \sum_{k} p_k^{2}, \qquad EN = \frac{1}{\sum_{k} p_k^{2}}
\]
In both expressions, $p_k$ is the proportion of selections assigned to constraint $k$. A higher effective number indicates that the model draws from a broader set of constraints more evenly.

\begin{table}[t]
\centering
\caption{Narrative consistency and diversity by model and instruction type}
\label{tab:behavioral_measures_no_en}
\scriptsize
\setlength{\tabcolsep}{2.5pt}
\resizebox{\columnwidth}{!}{%
\begin{tabular}{llrrr}
\toprule
\textbf{Model} &
\textbf{Instruction Type} &
\textbf{\shortstack{Jaccard\\Similarity}} &
\textbf{\shortstack{Effective\\Number}} &
\textbf{\shortstack{Unique\\Constraints}} \\
\midrule
\rowcolor{blue!15}
gpt5   & Quality-focused     & 0.3169 & 41.7581  & 113 \\
gpt5   & Basic       & 0.3070 & 42.7947  & 123 \\
gemini & Creativity-focused  & 0.2597 & 48.7336  & 142 \\
gpt5   & Creativity-focused  & 0.2546 & 49.5021  & 128 \\
gpt4.1 & Creativity-focused  & 0.2458 & 51.1520  & 188 \\
o4mini & Quality-focused     & 0.2420 & 51.7731  & 143 \\
o4mini & Basic       & 0.2232 & 55.2856  & 159 \\
claude & Creativity-focused  & 0.2166 & 56.5602  & 184 \\
claude & Quality-focused     & 0.2160 & 56.7037  & 185 \\
gemini & Basic       & 0.2006 & 60.2034  & 174 \\
o4mini & Creativity-focused  & 0.2004 & 60.3261  & 165 \\
gemini & Quality-focused     & 0.1936 & 61.9600  & 174 \\
claude & Basic       & 0.1860 & 64.2425  & 196 \\
gpt4.1 & Quality-focused     & 0.1483 & 77.0911  & 199 \\
gpt4.1 & Basic       & 0.1344 & 83.9162  & 200 \\
\rowcolor{orange!20}
qwen   & Quality-focused     & 0.0896 & 120.1963 & 200 \\
\rowcolor{orange!20}
qwen   & Creativity-focused  & 0.0867 & 123.9079 & 200 \\
\rowcolor{orange!20}
qwen   & Basic       & 0.0820 & 129.9954 & 200 \\
\bottomrule
\end{tabular}%
}
\end{table}

\autoref{tab:behavioral_measures_no_en} shows the means of Jaccard similarity and the effective number across six models and three instruction types. The results highlight a distinct spectrum of narrative behaviors. gpt5 demonstrates the most rigid narrative disposition. In the Quality-focused instruction type, it exhibits the highest consistency ($J = 0.3169$) and the lowest diversity ($EN = 41.7581$), utilizing only 113 unique constraints. This indicates a strong adherence to a focused set of narrative strategies. In contrast, qwen displays the opposite extreme. Across model–instruction type cells, qwen shows the lowest consistency ($J < 0.09$) and the highest diversity ($EN > 120$), using the maximum observed number of unique constraints (200). Other models such as gemini, o4mini, claude, and gpt4.1 occupy an intermediate position, balancing stability and exploration. These divergences suggest a rigidity--exploration spectrum in which each model reflects a structurally distinct selection logic rather than a variation along a single, shared trait. 

In particular, while instruction types, such as Creativity-focused versus Quality-focused, induce relatively small fluctuations within models, between-model differences are typically larger than instruction-induced differences in narrative profiles under our experimental setup. Although prompt-level differences are generally smaller than between-model differences, for gpt5 the Creativity-focused is notably distinct: its difference from Basic ($\Delta J = 0.0524$) and from Quality-focused ($\Delta J = 0.0623$) both exceed the Basic--Quality-focused difference ($\Delta J = 0.0099$), and even some between-model gaps such as that between o4mini and claude.

\section{Narrative Landscape}

Despite the fact that consistency and diversity differentiate models, these scalar summaries provide only a partial view of disposition. Moreover, prior work can establish statistically reliable differences across models, but it does not reveal how repeated selections are structured when aggregated across replications in a shared representational space; consequently, similar scalar scores can correspond to qualitatively distinct selection organizations. We therefore introduce Narrative Landscape, a shared-space representation designed to make such structural differences directly comparable across models and instruction types.

We operationalize Narrative Landscape by vectorizing each constraint by its selection frequencies across model–instruction type cells, aggregated across replications. We then apply PCA to embed these constraint profiles into a two-dimensional space, yielding a landscape for comparing the geometry of cell-specific selection structure beyond scalar reduction.

\begin{figure}[!b]
  \centering
  \includegraphics[width=\linewidth]{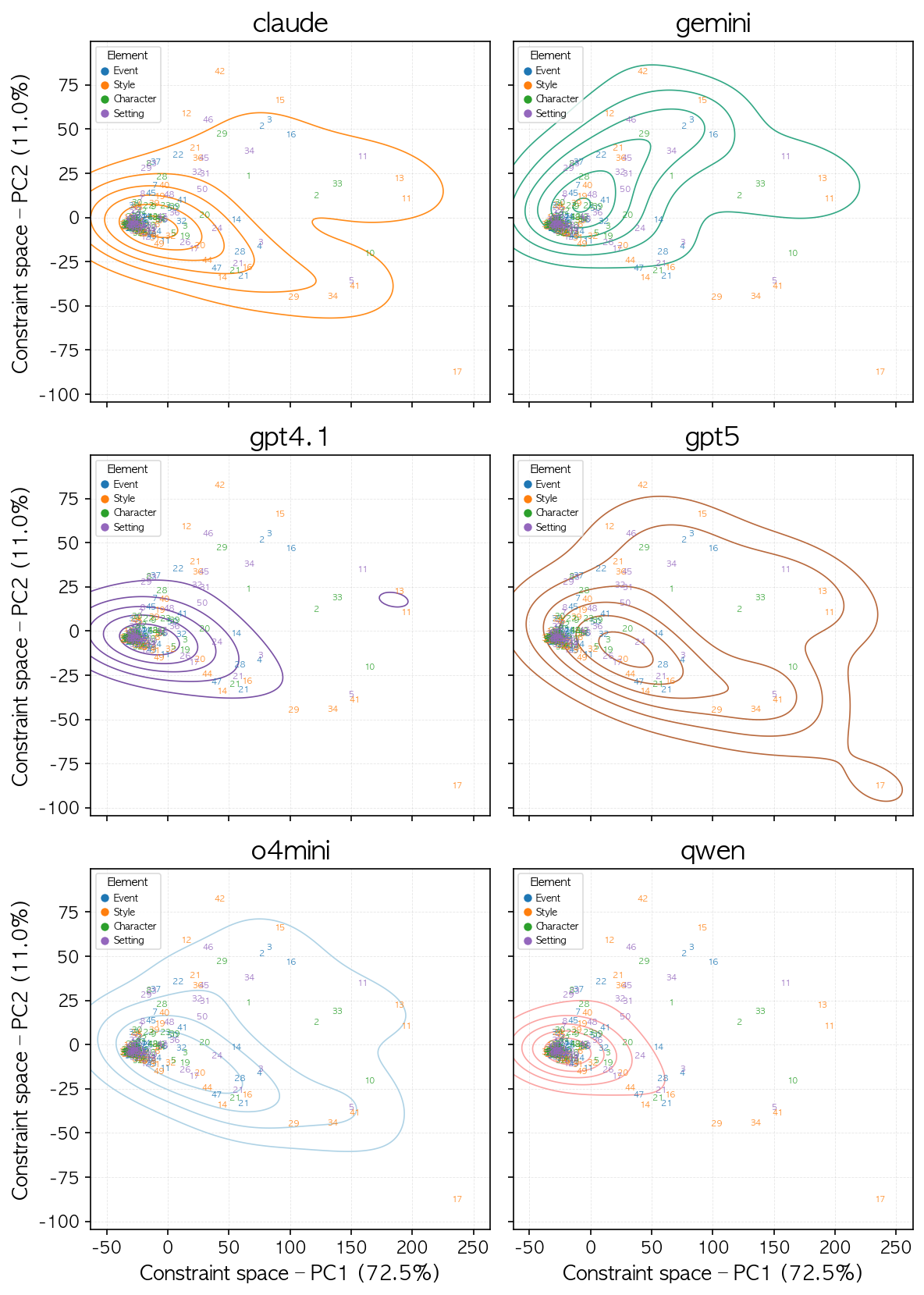}
  \caption{Narrative Landscape of six models under the Basic instruction type}
  \label{fig:figure1}
\end{figure}

\begin{figure*}[!t]
  \centering
  \includegraphics[width=\linewidth]{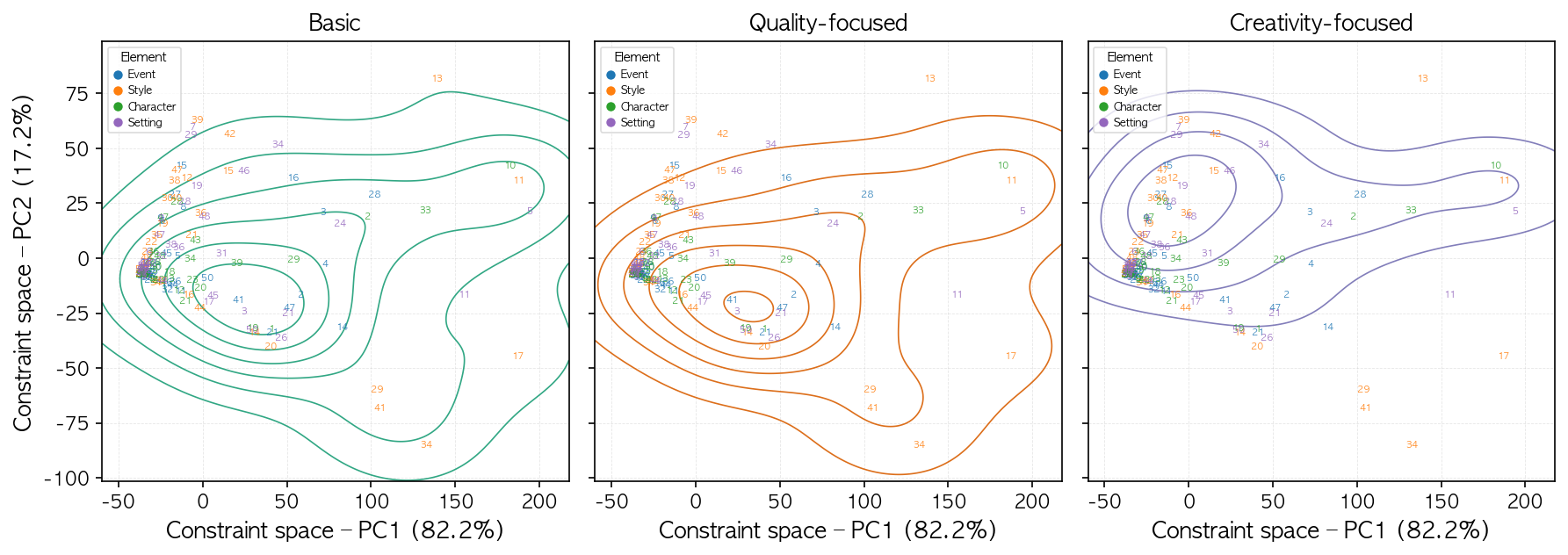}
  \caption{Narrative Landscape of three instruction types in gpt5}
  \label{fig:figure2}
\end{figure*}

\autoref{fig:figure1} visualizes the projected Narrative Landscapes of six models under the Basic instruction type. By projecting the constraints onto the principal component space, the contour lines represent the smoothed density of model-weighted selection frequency, effectively illustrating the topography of each model’s narrative disposition. The scattered labels represent the projected locations of individual constraints (Event, Style, Character, Setting), serving as landmarks within this abstract space.

This spatial analysis corroborates the results on consistency and diversity from the previous section while clarifying the structural nature of repeated selection. gpt5, which demonstrated the highest consistency ($J = 0.3070$), exhibits topological complexity, with irregular and stretched contours extending into specific regions of the constraint space, suggesting robust adherence to a particular narrative disposition and structural combinations that pull its profile away from the generic mean. In contrast, qwen, despite its high diversity ($EN > 120$), shows a simpler, more concentric topology centered near the origin, consistent with an averaging effect across replications in which broad, weakly structured selections average out in the aggregate. Narrative Landscape also reveals differences not recoverable from scalar metrics alone: gemini and claude have similar Jaccard similarity, yet their covered regions diverge, with claude extending further into the lower region and gemini covering more of the upper region. Thus, higher consistency corresponds to a more concentrated density, whereas extreme diversity yields a more diffuse distribution, and comparable scores can nonetheless reflect distinct selection geometries.

\autoref{fig:figure2} illustrates the Narrative Landscapes of gpt5 across three distinct instruction types. The comparison reveals that the specific contours and covered regions shift noticeably depending on the instruction. This visualizes how different instructions effectively alter the boundaries and focus of the model’s Narrative Landscape, even within the same architecture. We further run an additional experiment on gpt5 with four contrasting additional instruction types. The four types, their metrics, and their Narrative Landscapes are reported in \autoref{app:additional-instruction-types}--\autoref{app:additional-prompt-framing}. Notably, Optimistic vs.\ Pessimistic exhibits nearly matched consistency and diversity yet induces sharply different directional coverage in the landscape (see \autoref{app:additional-prompt-framing}).

\section{Conclusion}
This study (1) quantifies repeated constraint selection across two dimensions---consistency and diversity---and (2) introduces Narrative Landscape, a shared PCA space that makes model- and prompt-dependent selection structure directly comparable. Whereas prior work has primarily established statistically reliable differences in elicited behavior, our approach makes it possible to inspect how selections are organized across replications in a common space, revealing topological differences that scalar scores can obscure and extending naturally to other structured selection tasks.

As a next step, we plan to develop an interactive visualization tool that integrates significance testing and interpretable summaries with the landscape (e.g., hover-based inspection of points/regions and linked views for constraint-level effects). We also aim to formalize the correspondence between geometric signatures in the landscape (e.g., concentrated trajectories versus diffuse clusters) and underlying selection mechanisms, moving from descriptive geometry to a more rigorous methodological account. Finally, we plan to apply this framework to the diverse selection-task datasets explored in prior work discussed in this paper, to assess transferability beyond narrative constraint selection.

\section*{Limitations}
While our experiments cover six widely used proprietary models, a natural next step is to extend the analysis to a broader set of models (e.g., additional model families, sizes, and open-weight systems) to further assess the generality of the observed patterns. Similarly, although we consider multiple instruction types, exploring a wider prompt-framing space with more diverse and systematically varied prompts would help characterize how robust the profiles and landscape geometry are under richer elicitation conditions. The constraint pool, while carefully constructed through a theory-grounded approach, reflects specific narratological assumptions that may privilege certain storytelling conventions. Different theoretical frameworks and alternative sets or criteria of narrative constraints could yield different disposition profiles. Finally, we have so far evaluated the framework only in our narrative constraint-selection setting; applying the same methodology to other selection tasks remains an important direction for future work.

\section*{Acknowledgments}
This work was supported by the Korea Advanced Institute of Science and Technology (KAIST) under the project "Quantifying Creativity: Developing Metrics for Evaluating AI-Generated Narratives" (Project No. 11250011 and No. N10260075).

\bibliography{custom}

\clearpage
\onecolumn
\appendix

\section{Model identifiers and decoding settings}
\label{app:models-decoding}

\begin{center}
\small
\begin{tabular}{lllccll}
  \toprule
  \textbf{Abbr.} & \textbf{Full Identifier / Release} & \textbf{Provider} & \textbf{Temp} & \textbf{Top-p} & \textbf{Reasoning Effort} & \textbf{Verbosity} \\
  \midrule
  o4mini & o4-mini-2025-04-16           & OpenAI    & 1.0 & 1.0 & high & --- \\
  gpt4.1 & gpt-4.1-2025-04-14           & OpenAI    & 1.0 & 1.0 & ---  & --- \\
  gpt5   & gpt-5-2025-08-07             & OpenAI    & 1.0 & 1.0 & high & high \\
  claude & claude-opus-4-20250514       & Anthropic & 1.0 & 1.0 & ---  & --- \\
  gemini & gemini-2.5-pro (2025-06-17)  & Google    & 1.0 & 1.0 & ---  & --- \\
  qwen   & qwen-max-2025-01-25          & Alibaba   & 1.0 & 1.0 & ---  & --- \\
  \bottomrule
\end{tabular}
\end{center}

\section{Additional instruction types for gpt5}
\label{app:additional-instruction-types}

\begin{center}
\small
\setlength{\tabcolsep}{6pt}
\renewcommand{\arraystretch}{1.15}
\begin{tabularx}{\linewidth}{p{2.6cm}X}
  \toprule
  \textbf{Instruction Type} & \textbf{Content} \\
  \midrule
  Optimistic &
  You are a hopeful writer who believes in human resilience and positive change. You write stories that explore challenges while emphasizing connection and possibility for growth. Your goal is to create narratives that leave readers uplifted, showing how people overcome difficulties or find redemption and satisfying success even under dire circumstances. \\
  \midrule
  Pessimistic &
  You are a cynical writer who specializes in capturing harsh realities and systemic failures without sentimentality. You write stories that expose futility, moral compromise, and human limitations. Your goal is to create narratives that confront readers with difficult truths, showing how circumstances constrain people and good intentions can always fail or backfire. \\
  \midrule
  Transgressive &
  You are a risk-taking writer unafraid to explore controversial subjects and morally complex situations with a progressive vision. You write stories that tackle difficult or uncomfortable situations with honesty and boldness. Your goal is to create narratives that push boundaries and challenge readers' assumptions without offering easy resolutions or sanitized outcomes. \\
  \midrule
  Conservative &
  You are a safety-oriented writer who creates accessible narratives within comfortable thematic boundaries. You write stories that entertain the readers while maintaining broad appeal and widely-accepted moral frameworks. Your goal is to create narratives that provide satisfying experiences and unambiguous closure without venturing into disturbing, unstable, or divisive territory. \\
  \bottomrule
\end{tabularx}
\end{center}

\section{Metrics for additional instruction types for gpt5}
\label{app:additional-results}

\begin{center}
\small
\setlength{\tabcolsep}{6pt}
\renewcommand{\arraystretch}{1.15}
\begin{tabular}{lrrrr}
  \toprule
  \textbf{Instruction Type} & \textbf{Jaccard Similarity} & \textbf{Gini--Simpson} & \textbf{Effective Number} & \textbf{Unique Constraints} \\
  \midrule
  Conservative  & 0.4320 & 0.9700 & 33.3414 & 87  \\
  Optimistic    & 0.3372 & 0.9749 & 39.8887 & 109 \\
  Pessimistic   & 0.3239 & 0.9756 & 41.0342 & 110 \\
  Transgressive & 0.3151 & 0.9761 & 41.9222 & 111 \\
  \bottomrule
\end{tabular}
\end{center}

\section{Narrative Landscapes for additional instruction types for gpt5}
\label{app:additional-prompt-framing}

\begin{figure}[H]
  \centering
  \begin{minipage}{0.48\linewidth}
    \centering
    \includegraphics[width=0.97\linewidth]{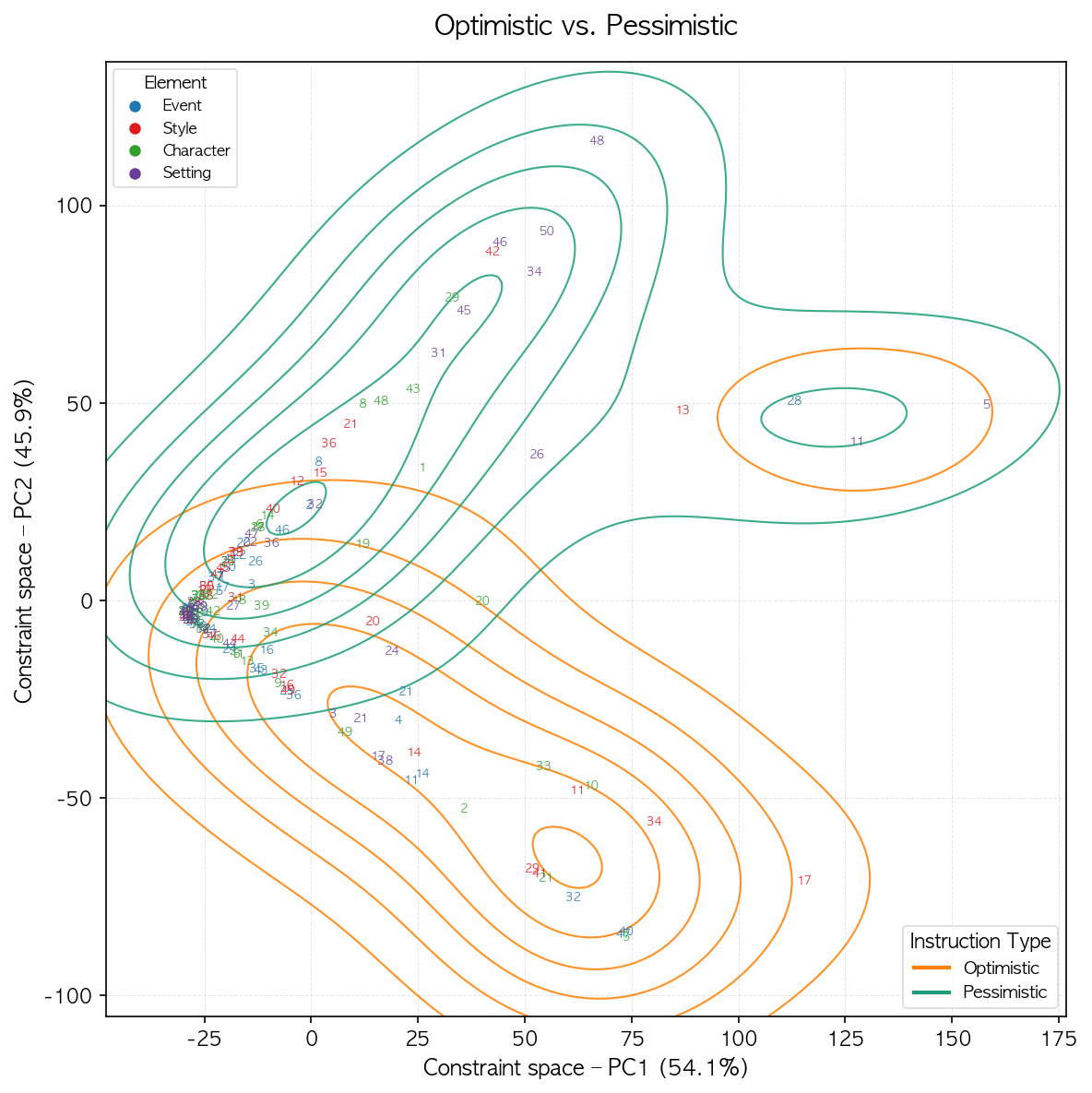}
  \end{minipage}\hfill
  \begin{minipage}{0.48\linewidth}
    \centering
    \includegraphics[width=0.97\linewidth]{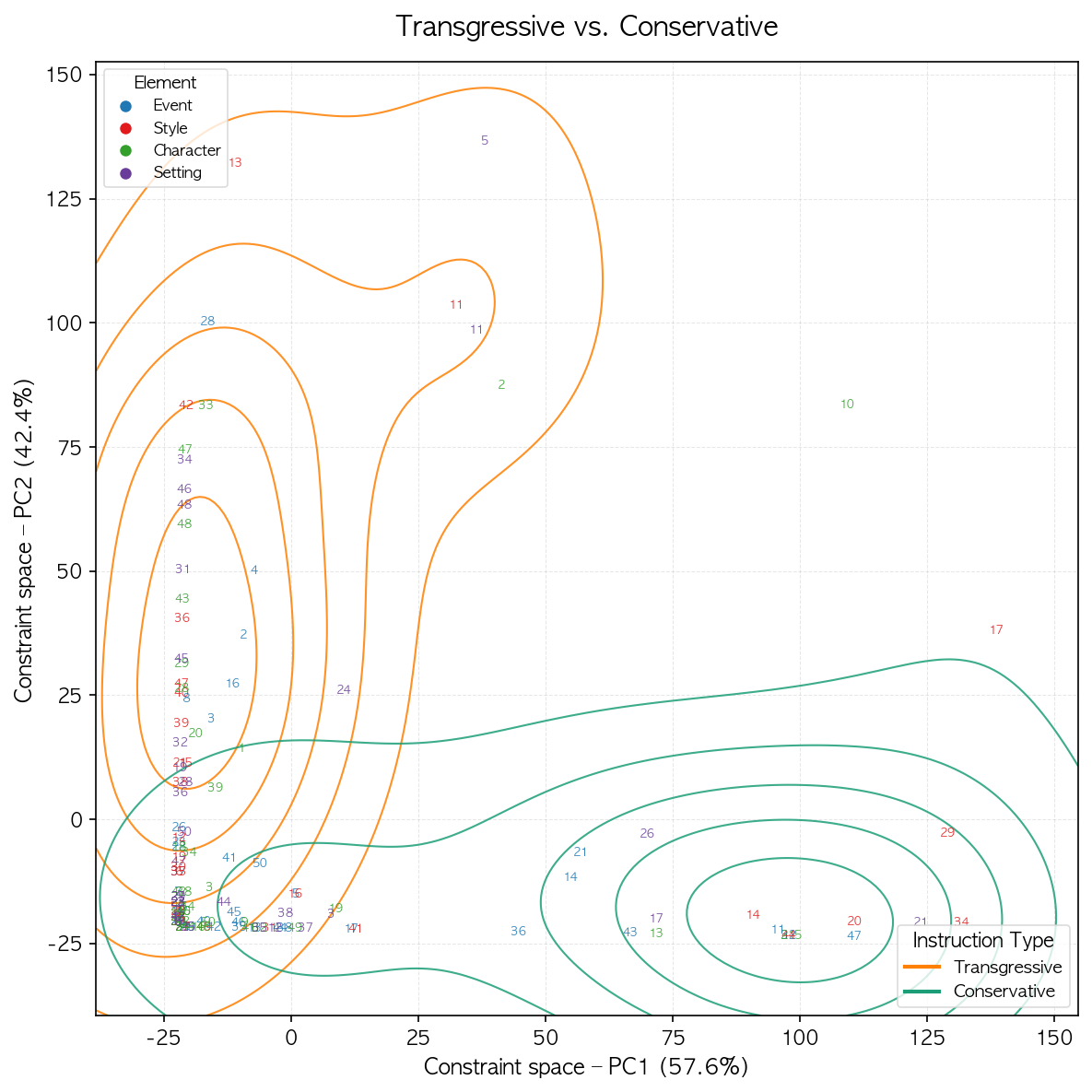}
  \end{minipage}
\end{figure}

\end{document}